\pdfoutput=1

\documentclass[11pt]{article}

\usepackage[]{latex/acl}

\usepackage{times}
\usepackage{latexsym}

\usepackage[T1]{fontenc}

\usepackage[utf8]{inputenc}

\usepackage{microtype}

\usepackage{inconsolata}

\usepackage{color}
\usepackage{amsmath}
\usepackage{amssymb}
\usepackage{booktabs}
\usepackage{graphicx} 
\usepackage{subcaption} 
\title{Sparse Contrastive Learning of Sentence Embeddings}

\author{Ruize An, Chen Zhang, Dawei Song \\
  Beijing Institute of Technology \\ 
  \texttt{\{rz.an,czhang,dwsong\}@bit.edu.cn} \\}

\begin{document}
\maketitle
\begin{abstract}
Recently, SimCSE has shown the feasibility of contrastive learning in training sentence embeddings and illustrates its expressiveness in spanning an aligned and uniform embedding space.
However, prior studies have shown that dense models could contain harmful parameters that affect the model performance, and it is no wonder that SimCSE can as well be invented with such parameters.
Driven by this, parameter sparsification is applied, where alignment and uniformity scores are used to measure the contribution of each parameter to the overall quality of sentence embeddings. 
Drawing from a preliminary study, we consider parameters with minimal contributions to be detrimental, as their sparsification results in improved model performance.
To discuss the ubiquity of detrimental parameters and remove them, more experiments on the standard semantic textual similarity (STS) tasks and transfer learning tasks are conducted, and the results show that the proposed sparsified SimCSE (SparseCSE) has excellent performance in comparison with SimCSE. 
Furthermore, through in-depth analysis, we establish the validity and stability of our sparsification method, showcasing that the embedding space generated by SparseCSE exhibits improved alignment compared to that produced by SimCSE. Importantly, the uniformity yet remains uncompromised.
\end{abstract}

\section{Introduction}

\begin{figure}[ht]
\centering
    \includegraphics[width=0.35\textwidth]{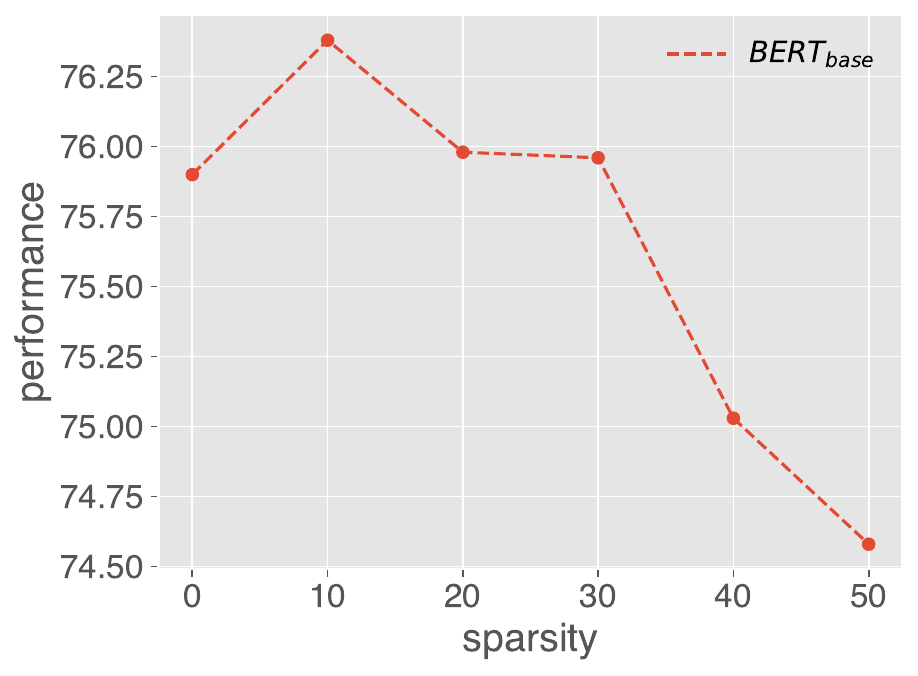}
    \caption{
    The average performance of SimCSE-BERT\textsubscript{base} on STS tasks when pruned at sparsity levels of 10\%, 20\%, 30\%, 40\% and 50\% respectively. Details of the pruning method can be found in Section~\ref{method}, while the task specifics and metrics are introduced in Section~\ref{Experiments}.}
    \label{fig:pre_peak}
\end{figure}

The task of learning universal sentence embeddings using large-scale pre-trained models has been extensively explored in prior research 
~\citep{
DBLP:conf/iclr/LogeswaranL18,
reimers-gurevych-2019-sentence,
DBLP:conf/emnlp/LiZHWYL20,
DBLP:conf/emnlp/ZhangHLLB20,
DBLP:conf/emnlp/GaoYC21,
DBLP:conf/emnlp/0001VKC21,
DBLP:conf/acl/YanLWZWX20,
DBLP:conf/acl/FengYCA022}. 
More recently, contrastive learning has been proposed as a method to enhance the quality of sentence embeddings 
~\citep{
DBLP:conf/wsdm/QiuHYW22,
DBLP:conf/emnlp/ZhangHLLB20,
DBLP:conf/emnlp/GaoYC21,
DBLP:conf/emnlp/0001VKC21,
DBLP:conf/acl/YanLWZWX20}. 
By employing contrastive learning, semantically similar sentences are brought closer together while dissimilar sentences are pushed apart, thereby a semantically-driven structure is established within the space of sentence embeddings.

Unsupervised SimCSE (unsup-SimCSE) is a notable framework for contrastive sentence embeddings~\citep{DBLP:conf/emnlp/GaoYC21}. It utilizes dropout as a simple data augmentation technique to create positive pairs and employs a cross-entropy objective based on cosine similarity for contrastive learning. 
Inspired by recent research on parameter sparsification~\citep{DBLP:conf/acl/XiaZC22, prasanna-etal-2020-bert,DBLP:conf/nips/HouHSJCL20,DBLP:conf/nips/MichelLN19}, particularly the works on the lottery ticket hypothesis (LTH)~\citep{DBLP:conf/iclr/FrankleC19,DBLP:conf/iclr/BaiWTL022,DBLP:conf/icml/FrankleD0C20,DBLP:conf/nlpcc/YangZWS22} showing its effectiveness in improving model performance through pruning, we hypothesize that certain parameters in SimCSE might hinder the representation of universal sentence embeddings. By removing these parameters, we anticipate an improvement in the model's performance.

To accurately estimate the contribution of each parameter, it is essential to consider properties that characterize contrastive representation learning. 
In the literature ~\citep{DBLP:conf/icml/0001I20}, two such properties have been proposed: alignment and uniformity.
Alignment measures the proximity of features derived from positive pairs, indicating how well the model captures semantic similarity. On the other hand, uniformity pertains to the distribution of features across the hypersphere, ensuring that the representations are spread out evenly. These properties offer valuable insights into understanding and evaluating contrastive representation learning.
Utilizing alignment and uniformity as guiding principles, we propose an innovative approach, named alignment and uniformity score, to quantify parameter contribution during the preparation phase for pruning.

Based on a pilot study presented in Figure \ref{fig:pre_peak}, we observed that model performance does not consistently decrease during pruning, instead it exhibits an upward trend when the model is less sparse. This suggests that the parameters with the lowest scores are detrimental to model performance, as evidenced by the performance improvement resulting from their pruning. Building upon this, we conducted a series of more extensive and detailed experiments to explore the ubiquity of detrimental parameters and assess the stability of our proposed pruning method.

Specifically, our approach consists of three stages: training, parameter sparsification, and rewinding. 
First, we train an unsupervised SimCSE model using a pre-trained language model (LM). Then, we estimate alignment and uniformity scores for each parameter based on the trained model's feedback. Parameters with low scores are pruned and varying sparsity is attempted in formal experiments than in pilot study to clearly identify harmful parameters. Finally, the remaining parameters are initialized, and the pruned model is fine-tuned to regain its performance.
Our model is thus named SparseCSE.

We extensively evaluate SparseCSE on seven STS tasks and seven transfer learning tasks. Results show that SparseCSE outperforms SimCSE, demonstrating its superior performance. Our pruning method is also shown to effectively identify the optimal sparsity for pruning, further enhancing performance.
Further analysis reveals the stability of our pruning method across multiple tasks. 
Comparison with other works highlights the similarity of SparseCSE to SimCSE in uniformity and its competitive performance in alignment.

\begin{figure}[ht]
\centering
\includegraphics[width=0.45\textwidth]{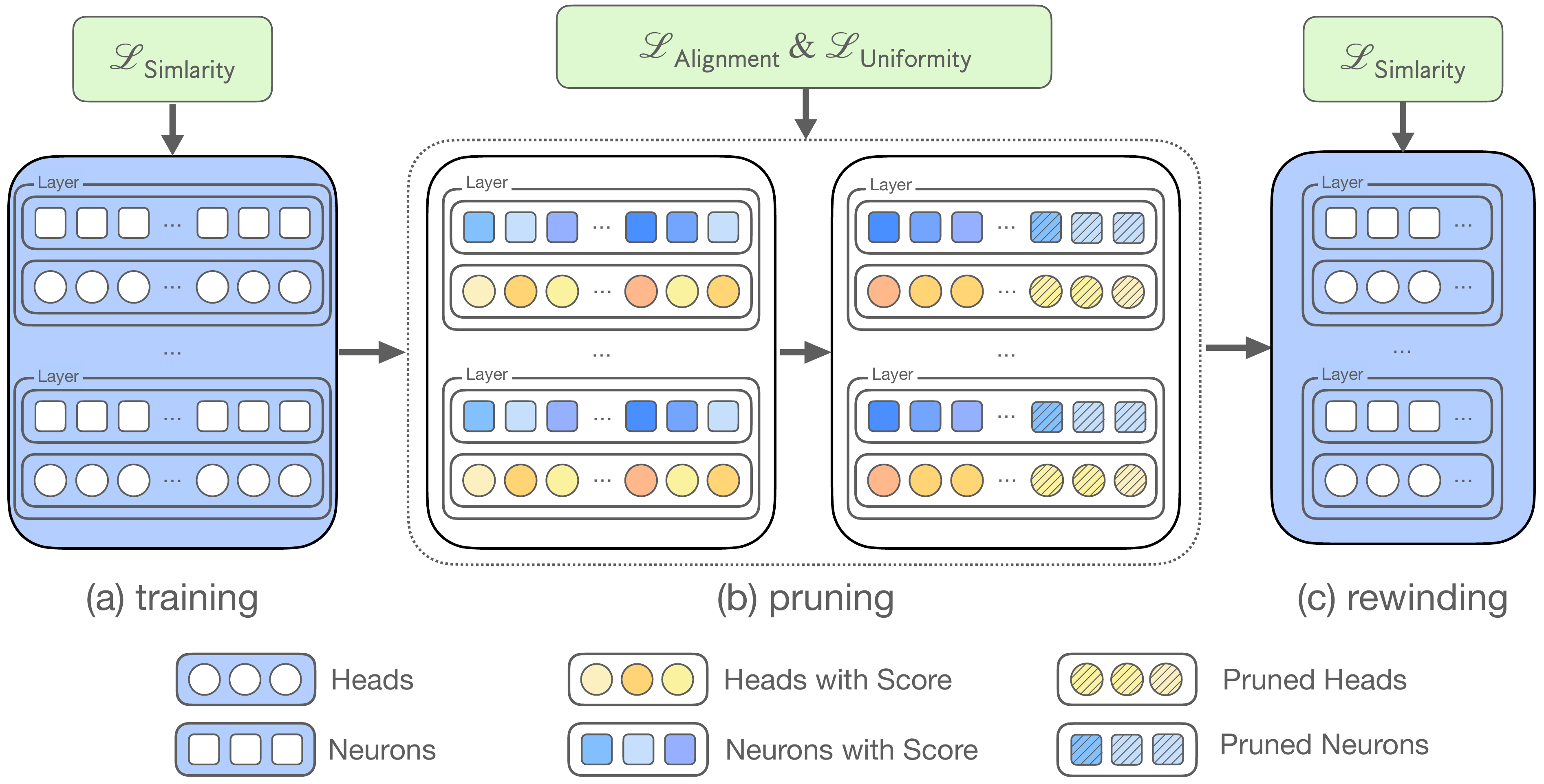}
    \caption{The process of obtaining SparseCSE}
    \label{fig:sparse}
\end{figure}

\section{Our Method} 
\label{method}

Similar to the lottery ticket approach~\citep{DBLP:conf/iclr/FrankleC19}, our method is illustrated in Figure \ref{fig:sparse}, following a training, pruning, and rewinding paradigm.

\subsection{Training and Rewinding}

To effectively train a model that captures universal sentence embeddings, we adopt a contrastive framework similar to the previous work ~\citep{DBLP:conf/emnlp/GaoYC21}. This framework is also utilized during the rewinding stage. In this framework, we employ dropout to create positive representation pairs $(h_i, h_i^+)$ for each sentence $x_i$ in a collection of sentences ${x_i}_{i=1}^m$. The training objective for this contrastive framework, using a mini-batch of $N$ pairs, can be expressed as follows:
\begin{equation}\nonumber
\begin{aligned}
\mathcal{L}_{\sf similarity}^{(i)} = -
\log 
\dfrac
{e^{{\mathbf{sim} ({h_i} , {h_i^+}) } / {\tau}}}
{\sum_{j=1}^n e^{{\mathbf{sim} ( {h_i} , {h_j^+} ) } / {\tau}}},
\end{aligned}
\end{equation}
where $\tau$ is a temperature hyperparameter and $\mathbf{sim}({h_1}, {h_2})$ represents the cosine similarity $\dfrac{{h_1}^\mathsf{T}\cdot{h_2}}{\Vert{h_1}\Vert\cdot\Vert{h_2}\Vert}$.

During training, an initial pretrained language model (LM) is utilized, and all parameters are involved in this phase. However, during rewinding, only the remaining parameters after pruning are applied to the LM, with their values initialized to their early-stage pre-training values. The objective of rewinding is to enable the pruned model to restore its performance prior to pruning.

\subsection{Pruning}

BERT~\citep{devlin-etal-2019-bert} (or Roberta~\citep{DBLP:journals/corr/abs-1907-11692}) is composed of multiple stacked encoder layers known as transformers. 
Each transformer encoder consists of a multi-head self-attention block (MHA) and a feed-forward network block (FFN). 
In line with prior research~\citep{prasanna-etal-2020-bert,DBLP:conf/nips/HouHSJCL20,DBLP:conf/nips/MichelLN19}, our pruning approach primarily focuses on sparsifying the attention heads in the MHA blocks and the intermediate neurons in the FFN blocks.
To determine which parameters to prune, we associate a set of mask variables with them~\citep{DBLP:conf/emnlp/YangZS22,DBLP:conf/nlpcc/YangZWS22} 
and compare the model's performance before and after the operation. 

For a MHA block with $N_H$ independent heads, the $i$-th head is parameterized by $\mathbf{W}_{Q}^{(i)}$, $\mathbf{W}_{K}^{(i)}$, $\mathbf{W}_{V}^{(i)}\in\mathbb{R}^{d\times d_{A}}$, and $\mathbf{W}_O^{(i)}\in\mathbb{R}^{d_{A}\times d}$, all parallel heads are further summed to produce the final output. Then variable $\xi^{(i)}$ with values in $\{0, 1\}$ is defined for masking each attention head, and it can be represented as:

\begin{equation}\nonumber
\begin{aligned}
&{\rm MHA}(\mathbf{X})
=\sum_{i=1}^{N_H}\xi^{(i)}{\rm Attn}_{\mathbf{W}_{Q}^{(i)},\mathbf{W}_{K}^{(i)},\mathbf{W}_{V}^{(i)},\mathbf{W}_O^{(i)}}^{(i)}
(\mathbf{X}),
\end{aligned}
\end{equation}
where the input $\mathbf{X}\in\mathbb{R}^{l\times d}$ represents a $l$-length sequence of $d$-dimensional vectors and $\xi^{(i)}$ is designed as a switching value, when $\xi^{(i)}$ equal to 1, it means keeping the attention head retained, and when it equal to 0 means removing that attention head from the MHA.

On the other hand, a FFN block includes two fully-connected layers parameterized by $\mathbf{W}_1\in\mathbb{R}^{d\times D_{F}}$ and $\mathbf{W}_2\in\mathbb{R}^{D_{F}\times d}$, denoting $D_{F}$ as the number of neurons in the intermediate layer of FFN. 
Likewise, we define variable $\nu$ to mask the neurons in the intermediate layer of FFN: 
\begin{equation}\nonumber
{\rm FFN}(\mathbf{A})= 
\sum_{i=1}^{D_{F}}{\nu^{(i)}}{\rm GELU}_{\mathbf{W}_1, \mathbf{W}_2}(\mathbf{A}), 
\end{equation}
where the input $\mathbf{A}\in\mathbb{R}^{l\times d}$ defines a $d$-dimensional vectors with $l$-length sequence.

\subsection{Alignment and Uniformity Score}

In order to determine the parameters that have a greater impact on the distribution of universal sentence embeddings, we introduce a joint objective based on the alignment and uniformity properties~\citep{DBLP:conf/icml/0001I20}. 

Here is the formulation of the alignment loss:
\begin{equation}\nonumber
\begin{aligned}
\mathcal{L}_{\sf Alignment} \triangleq 
\log{
\mathop{\mathbb{E}}\limits_{{\mathbf{x_i}, \mathbf{x_i}^+}\sim \mathcal{N}_{pos} }
{{ \Vert \mathbf{h_i} - \mathbf{h_i}^+ \Vert}^2}
}, 
\end{aligned}
\end{equation}
where $h_i, {h_i}^+$ are representations of $x_i, {x_i}^+$, which are a pair of positive sentences in a batch of $N_{pos}$ sentences. It indicates that the sentences with similar semantics are expected to be closer in the embedding space.

And, here is the formulation of the uniformity loss:
\begin{equation}\nonumber
\begin{aligned}
\mathcal{L}_{\sf Uniformity} \triangleq 
\log{
\mathop{\mathbb{E}}\limits_{{\mathbf{x_i}, \mathbf{x_j}}\sim \mathcal{N} }
{{\rm e}^{{-2 \Vert \mathbf{h_i} - \mathbf{h_j} \Vert}^2}} 
}, 
\end{aligned}
\end{equation}
where $h_i, h_j$ are representations of $x_i, x_j$, which are different sentences in a batch of $N$ sentences. It indicates that sentence embeddings with different semantics are supposed to distribute on the hypersphere by larger distances.

To balance the alignment and uniformity, we introduce a coefficient $\lambda$ to quantify the tradeoff. The joint loss $\mathcal{L}_{\sf Score}$ for further score calculation can be be written as below:
\begin{equation}\nonumber
\begin{aligned}
\mathcal{L}_{\sf Score} = \lambda\cdot\mathcal{L}_{\sf Alignment}+
(1-\lambda)\cdot\mathcal{L}_{\sf Uniformity},
\end{aligned}
\end{equation}

Finally, according to the literature~\citep{DBLP:conf/iclr/MolchanovTKAK17}, the scores of the attention heads in MHA and the intermediate neurons in FFN can be depicted as:

\begin{equation}\nonumber
\begin{aligned}
\mathbb{I}_{\sf head}^{(i)}&=\mathbb{E}_{\mathcal{D}}\left|\frac{\partial\mathcal{L}_{\sf Score}}{\partial\xi^{(i)}}\right|,\\ 
\mathbb{I}_{\sf neuron}^{(i)}&=\mathbb{E}_{\mathcal{D}}\left|\frac{\partial\mathcal{L}_{\sf Score}}{\partial \nu^{(i)}}\right|,
\end{aligned}
\end{equation}
where $\mathcal{D}$ is a data distribution, $\mathbb{E}$ represents expectation.

After estimating the scores, we rank the attention heads and intermediate neurons respectively with the scores, and prune the parameters with low scores according to the constraint of the given sparsity.

\begin{table*}[ht]
\renewcommand{\arraystretch}{}
\resizebox{\textwidth}{!}{
\begin{tabular}{@{}ccccccccc@{}}
\toprule
\multicolumn{1}{l}{}    & STS12 & STS13 & STS14 & STS15 & STS16 & STS-B & SICK-R & Avg   \\ \midrule
SimCSE-BERT\textsubscript{base}   & 70.37 & 82.53 & 73.46 & 81.58 & 77.61 & 76.55 & 69.22  & 75.9  \\
SparseCSE\textsubscript{2\%} 
& $70.15^{\textcolor{red!70}{-\textbf{0.22}}}$ 
& $82.25^{\textcolor{red!70}{-\textbf{0.28}}}$ 
& $74.16^{\textcolor{red!70}{+\textbf{0.70}}}$ 
& $82.15^{\textcolor{red!70}{+\textbf{0.57}}}$ 
& $78.52^{\textcolor{red!70}{+\textbf{0.91}}}$ 
& $78.71^{\textcolor{red!70}{+\textbf{2.16}}}$ 
& $72.76^{\textcolor{red!70}{+\textbf{3.54}}}$  
& $76.96^{\textcolor{red!70}{+\textbf{1.06}}}$ \\

SparseCSE\textsubscript{best} 
& $71.70^{\textcolor{red!70}{+\textbf{1.33}}}_{10\%}$ 
& $83.41^{\textcolor{red!70}{+\textbf{0.88}}}_{25\%}$ 
& $74.16^{\textcolor{red!70}{+\textbf{0.70}}}_{2\%}$ 
& $82.58^{\textcolor{red!70}{+\textbf{1.00}}}_{25\%}$ 
& $79.10^{\textcolor{red!70}{+\textbf{1.49}}}_{4\%}$ 
& $78.71^{\textcolor{red!70}{+\textbf{2.16}}}_{2\%}$ 
& $72.76^{\textcolor{red!70}{+\textbf{3.54}}}_{2\%}$  
& $77.49^{\textcolor{red!70}{+\textbf{1.59}}}_{}$ \\
\midrule
SimCSE-BERT\textsubscript{large}                  
& 69.93 & 84.04 & 75.15 & 82.99 & 78.32 & 79.12 & 74.16  & 77.67 \\
SparseCSE\textsubscript{2\%}  
& $69.31^{\textcolor{red!70}{-\textbf{0.62}}}$ 
& $83.69^{\textcolor{red!70}{-\textbf{0.35}}}$ 
& $75.72^{\textcolor{red!70}{+\textbf{0.57}}}$  
& $83.21^{\textcolor{red!70}{+\textbf{0.22}}}$  
& $79.34^{\textcolor{red!70}{+\textbf{1.02}}}$  
& $79.41^{\textcolor{red!70}{+\textbf{0.29}}}$   
& $74.76^{\textcolor{red!70}{+\textbf{0.60}}}$   
& $77.92^{\textcolor{red!70}{+\textbf{0.25}}}$  \\
SparseCSE\textsubscript{best}  
& $70.67^{\textcolor{red!70}{+\textbf{0.74}}}_{1\%}$ 
& $84.60^{\textcolor{red!70}{+\textbf{0.56}}}_{8\%}$ 
& $75.84^{\textcolor{red!70}{+\textbf{0.69}}}_{8\%}$  
& $83.21^{\textcolor{red!70}{+\textbf{0.22}}}_{1\%}$  
& $79.60^{\textcolor{red!70}{+\textbf{1.28}}}_{8\%}$  
& $79.41^{\textcolor{red!70}{+\textbf{0.29}}}_{1\%}$   
& $75.27^{\textcolor{red!70}{+\textbf{1.11}}}_{3\%}$   
& $78.32^{\textcolor{red!70}{+\textbf{0.64}}}_{}$  \\

\midrule
SimCSE-Roberta\textsubscript{base}                  & 67.45 & 81.28 & 72.74 & 81.31 & 80.87 & 80.12 & 68.37  & 76.02 \\
SparseCSE\textsubscript{1\%}
& $67.85^{\textcolor{red!70}{+\textbf{0.40}}}$  
& $81.32^{\textcolor{red!70}{+\textbf{0.04}}}$  
& $73.09^{\textcolor{red!70}{+\textbf{0.35}}}$  
& $81.82^{\textcolor{red!70}{+\textbf{0.51}}}$  
& $81.02^{\textcolor{red!70}{+\textbf{0.15}}}$   
& $80.29^{\textcolor{red!70}{+\textbf{0.17}}}$  
& $68.76^{\textcolor{red!70}{+\textbf{0.39}}}$   
& $76.31^{\textcolor{red!70}{+\textbf{0.29}}}$  \\ 
SparseCSE\textsubscript{best}  
& $68.05^{\textcolor{red!70}{+\textbf{0.60}}}_{4\%}$ 
& $81.82^{\textcolor{red!70}{+\textbf{0.54}}}_{4\%}$ 
& $73.32^{\textcolor{red!70}{+\textbf{0.58}}}_{4\%}$  
& $82.29^{\textcolor{red!70}{+\textbf{0.98}}}_{20\%}$  
& $81.02^{\textcolor{red!70}{+\textbf{0.15}}}_{2\%}$  
& $80.29^{\textcolor{red!70}{+\textbf{0.17}}}_{1\%}$   
& $68.76^{\textcolor{red!70}{+\textbf{0.39}}}_{1\%}$   
& $76.48^{\textcolor{red!70}{+\textbf{0.46}}}_{}$  \\

\bottomrule
\end{tabular}
}
\caption{
Performance of sparseCSE on STS tasks. Each backbone has three rows: the baseline, the result with optimal sparsity based on average score, and the result with optimal sparsity based on each task. The optimal sparsity values are shown in the bottom right corner. The improvements over the baseline are highlighted in red in the upper right corner.}
\label{tab:sts_result}
\end{table*}

\section{Experiments}
\label{Experiments}
\subsection{Baselines \& Implementation}
We start by training unsup-SimCSE models using popular language models (BERT\textsubscript{base}, BERT\textsubscript{large}, Roberta\textsubscript{base}) as our baselines. 
Both training and rewinding process of sparseCSE follow the training details of SimCSE~\citep{DBLP:conf/emnlp/GaoYC21}. Pruning process is produced with varying sparsity levels from 1\% to 50\% and different value of coefficient $\lambda$.
More details are shown in Appendix \ref{Pruning}.

\subsection{Evaluation}

Following SimCSE~\citep{DBLP:conf/emnlp/GaoYC21}, we evaluate sentence embeddings on 7 semantic textual similarity (STS) tasks, which include STS 2012–2016~\citep{agirre-etal-2012-semeval,agirre-etal-2013-sem,agirre-etal-2014-semeval,agirre-etal-2015-semeval,agirre-etal-2016-semeval}, STS Benchmark~\citep{cer-etal-2017-semeval} and SICK-Relatedness~\citep{DBLP:conf/lrec/MarelliMBBBZ14}. 
STS tasks can reveal the ability of clustering semantically similar sentences, which is one of the main goals for sentence embeddings. 
Furthermore, we also introduce 7 transfer learning tasks into evaluation as a supplementary prove, and the details of the tasks are shown in Appendix~\ref{Transfer}.

\subsection{Main Results}
Table~\ref{tab:sts_result} shows the results on STS tasks. 
The best results based on each task are all improved, and the model on BERT\textsubscript{base} improves the average result from 75.9\% to 77.49\%. We also determine an optimal sparsity corresponding to the best average score of all tasks. 
We observe that pruning the models with this specific sparsity level leads to improvements in almost every task. 
The results of transfer task are shown in Table~\ref{tab:transfer} in Appendix~\ref{Transfer}, where the same trend prove the ubiquity of the phenomenon found in Table~\ref{tab:sts_result}.

Considering the results comprehensively, we observe that the pruned models tend to exhibit optimal performance at lower sparsity levels. 
In order to carry out an in-depth analysis of this phenomenon, a detailed discussion on varying sparsity and tradeoff of alignment and uniformity are produced in Appendix~\ref{6.3} and~\ref{6.4}.


\begin{figure}[ht]
\centering
    \includegraphics[width=0.38\textwidth]{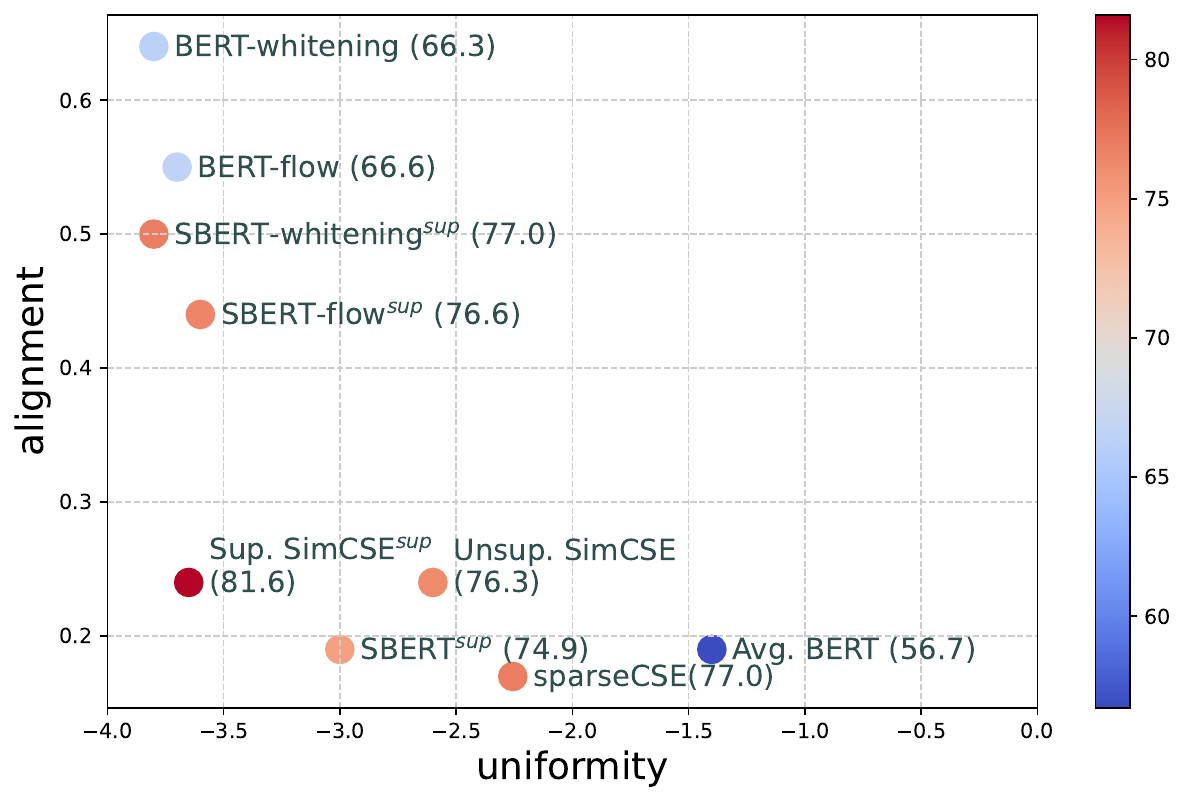}
    \caption{Analysis on alignment and uniformity (the smaller, the better). Points represent average STS performance using BERT\textsubscript{base}, with "sup" marked of supervised methods.}
    \label{fig:alignment_vs_uniformity}
\end{figure}

\paragraph{Evaluation on Alignment and Uniformity} 

Figure~\ref{fig:alignment_vs_uniformity} illustrates the uniformity and alignment scores of various methods along with their performance on the STS task. The methods include BERT~\citep{devlin-etal-2019-bert}, SimCSE~\citep{DBLP:conf/emnlp/GaoYC21}, SBERT~\citep{reimers-gurevych-2019-sentence}, BERT-flow~\citep{li-etal-2020-sentence}, BERT-whitening~\citep{DBLP:journals/corr/abs-2103-15316} and sparseCSE.
As a sparse version of Unsupervised SimCSE, sparseCSE inherits its advantages in alignment compared to post-processing methods (like BERT-flow and BERT-whitening) and uniformity compared to pre-trained embeddings (like BERT). Benefited from sparsity based on both alignment and uniformity properties, sparseCSE demonstrates significant improvements in alignment compared to the state-of-the-art models including the original model and some supervised models (like SBERT and supervised SimCSE), while achieving comparable uniformity scores.

\section{Conclusions}
In conclusion, this paper introduces a parameter sparsification technique based on alignment and uniformity scores, resulting in the development of SparseCSE, which exhibits impressive performance. 
Through extensive evaluation on STS tasks, transfer learning tasks, and comparison in terms of alignment and uniformity, SparseCSE demonstrates its competitive edge in sentence embedding.

\nocite{*}
\clearpage

\bibliography{custom}

\begin{thebibliography}{45}
\expandafter\ifx\csname natexlab\endcsname\relax\def\natexlab#1{#1}\fi

\bibitem[{Agirre et~al.(2015)Agirre, Banea, Cardie, Cer, Diab, Gonzalez-Agirre, Guo, Lopez-Gazpio, Maritxalar, Mihalcea, Rigau, Uria, and Wiebe}]{agirre-etal-2015-semeval}
Eneko Agirre, Carmen Banea, Claire Cardie, Daniel Cer, Mona Diab, Aitor Gonzalez-Agirre, Weiwei Guo, I{\~n}igo Lopez-Gazpio, Montse Maritxalar, Rada Mihalcea, German Rigau, Larraitz Uria, and Janyce Wiebe. 2015.
\newblock \href {https://doi.org/10.18653/v1/S15-2045} {{S}em{E}val-2015 task 2: Semantic textual similarity, {E}nglish, {S}panish and pilot on interpretability}.
\newblock In \emph{Proceedings of the 9th International Workshop on Semantic Evaluation ({S}em{E}val 2015)}, pages 252--263, Denver, Colorado. Association for Computational Linguistics.

\bibitem[{Agirre et~al.(2014)Agirre, Banea, Cardie, Cer, Diab, Gonzalez-Agirre, Guo, Mihalcea, Rigau, and Wiebe}]{agirre-etal-2014-semeval}
Eneko Agirre, Carmen Banea, Claire Cardie, Daniel Cer, Mona Diab, Aitor Gonzalez-Agirre, Weiwei Guo, Rada Mihalcea, German Rigau, and Janyce Wiebe. 2014.
\newblock \href {https://doi.org/10.3115/v1/S14-2010} {{S}em{E}val-2014 task 10: Multilingual semantic textual similarity}.
\newblock In \emph{Proceedings of the 8th International Workshop on Semantic Evaluation ({S}em{E}val 2014)}, pages 81--91, Dublin, Ireland. Association for Computational Linguistics.

\bibitem[{Agirre et~al.(2016)Agirre, Banea, Cer, Diab, Gonzalez-Agirre, Mihalcea, Rigau, and Wiebe}]{agirre-etal-2016-semeval}
Eneko Agirre, Carmen Banea, Daniel Cer, Mona Diab, Aitor Gonzalez-Agirre, Rada Mihalcea, German Rigau, and Janyce Wiebe. 2016.
\newblock \href {https://doi.org/10.18653/v1/S16-1081} {{S}em{E}val-2016 task 1: Semantic textual similarity, monolingual and cross-lingual evaluation}.
\newblock In \emph{Proceedings of the 10th International Workshop on Semantic Evaluation ({S}em{E}val-2016)}, pages 497--511, San Diego, California. Association for Computational Linguistics.

\bibitem[{Agirre et~al.(2012)Agirre, Cer, Diab, and Gonzalez-Agirre}]{agirre-etal-2012-semeval}
Eneko Agirre, Daniel Cer, Mona Diab, and Aitor Gonzalez-Agirre. 2012.
\newblock \href {https://aclanthology.org/S12-1051} {{S}em{E}val-2012 task 6: A pilot on semantic textual similarity}.
\newblock In \emph{*{SEM} 2012: The First Joint Conference on Lexical and Computational Semantics {--} Volume 1: Proceedings of the main conference and the shared task, and Volume 2: Proceedings of the Sixth International Workshop on Semantic Evaluation ({S}em{E}val 2012)}, pages 385--393, Montr{\'e}al, Canada. Association for Computational Linguistics.

\bibitem[{Agirre et~al.(2013)Agirre, Cer, Diab, Gonzalez-Agirre, and Guo}]{agirre-etal-2013-sem}
Eneko Agirre, Daniel Cer, Mona Diab, Aitor Gonzalez-Agirre, and Weiwei Guo. 2013.
\newblock \href {https://aclanthology.org/S13-1004} {*{SEM} 2013 shared task: Semantic textual similarity}.
\newblock In \emph{Second Joint Conference on Lexical and Computational Semantics (*{SEM}), Volume 1: Proceedings of the Main Conference and the Shared Task: Semantic Textual Similarity}, pages 32--43, Atlanta, Georgia, USA. Association for Computational Linguistics.

\bibitem[{Amplayo et~al.(2022)Amplayo, Brazinskas, Suhara, Wang, and Liu}]{DBLP:conf/sigir/AmplayoBS0L22}
Reinald~Kim Amplayo, Arthur Brazinskas, Yoshi Suhara, Xiaolan Wang, and Bing Liu. 2022.
\newblock \href {https://doi.org/10.1145/3477495.3532676} {Beyond opinion mining: Summarizing opinions of customer reviews}.
\newblock In \emph{{SIGIR} '22: The 45th International {ACM} {SIGIR} Conference on Research and Development in Information Retrieval, Madrid, Spain, July 11 - 15, 2022}, pages 3447--3450. {ACM}.

\bibitem[{Bai et~al.(2022)Bai, Wang, Tao, Li, and Fu}]{DBLP:conf/iclr/BaiWTL022}
Yue Bai, Huan Wang, Zhiqiang Tao, Kunpeng Li, and Yun Fu. 2022.
\newblock \href {https://openreview.net/forum?id=fOsN52jn25l} {Dual lottery ticket hypothesis}.
\newblock In \emph{The Tenth International Conference on Learning Representations, {ICLR} 2022, Virtual Event, April 25-29, 2022}. OpenReview.net.

\bibitem[{Cer et~al.(2017)Cer, Diab, Agirre, Lopez-Gazpio, and Specia}]{cer-etal-2017-semeval}
Daniel Cer, Mona Diab, Eneko Agirre, I{\~n}igo Lopez-Gazpio, and Lucia Specia. 2017.
\newblock \href {https://doi.org/10.18653/v1/S17-2001} {{S}em{E}val-2017 task 1: Semantic textual similarity multilingual and crosslingual focused evaluation}.
\newblock In \emph{Proceedings of the 11th International Workshop on Semantic Evaluation ({S}em{E}val-2017)}, pages 1--14, Vancouver, Canada. Association for Computational Linguistics.

\bibitem[{Conneau and Kiela(2018)}]{conneau2018senteval}
Alexis Conneau and Douwe Kiela. 2018.
\newblock Senteval: An evaluation toolkit for universal sentence representations.
\newblock \emph{arXiv preprint arXiv:1803.05449}.

\bibitem[{Conneau et~al.(2017)Conneau, Kiela, Schwenk, Barrault, and Bordes}]{conneau-etal-2017-supervised}
Alexis Conneau, Douwe Kiela, Holger Schwenk, Lo{\"\i}c Barrault, and Antoine Bordes. 2017.
\newblock \href {https://doi.org/10.18653/v1/D17-1070} {Supervised learning of universal sentence representations from natural language inference data}.
\newblock In \emph{Proceedings of the 2017 Conference on Empirical Methods in Natural Language Processing}, pages 670--680, Copenhagen, Denmark. Association for Computational Linguistics.

\bibitem[{Devlin et~al.(2019)Devlin, Chang, Lee, and Toutanova}]{devlin-etal-2019-bert}
Jacob Devlin, Ming-Wei Chang, Kenton Lee, and Kristina Toutanova. 2019.
\newblock \href {https://doi.org/10.18653/v1/N19-1423} {{BERT}: Pre-training of deep bidirectional transformers for language understanding}.
\newblock In \emph{Proceedings of the 2019 Conference of the North {A}merican Chapter of the Association for Computational Linguistics: Human Language Technologies, Volume 1 (Long and Short Papers)}, pages 4171--4186, Minneapolis, Minnesota. Association for Computational Linguistics.

\bibitem[{Dolan and Brockett(2005)}]{dolan-brockett-2005-automatically}
William~B. Dolan and Chris Brockett. 2005.
\newblock \href {https://aclanthology.org/I05-5002} {Automatically constructing a corpus of sentential paraphrases}.
\newblock In \emph{Proceedings of the Third International Workshop on Paraphrasing ({IWP}2005)}.

\bibitem[{Feng et~al.(2022)Feng, Yang, Cer, Arivazhagan, and Wang}]{DBLP:conf/acl/FengYCA022}
Fangxiaoyu Feng, Yinfei Yang, Daniel Cer, Naveen Arivazhagan, and Wei Wang. 2022.
\newblock \href {https://doi.org/10.18653/v1/2022.acl-long.62} {Language-agnostic {BERT} sentence embedding}.
\newblock In \emph{Proceedings of the 60th Annual Meeting of the Association for Computational Linguistics (Volume 1: Long Papers), {ACL} 2022, Dublin, Ireland, May 22-27, 2022}, pages 878--891. Association for Computational Linguistics.

\bibitem[{Frankle and Carbin(2019)}]{DBLP:conf/iclr/FrankleC19}
Jonathan Frankle and Michael Carbin. 2019.
\newblock \href {https://openreview.net/forum?id=rJl-b3RcF7} {The lottery ticket hypothesis: Finding sparse, trainable neural networks}.
\newblock In \emph{7th International Conference on Learning Representations, {ICLR} 2019, New Orleans, LA, USA, May 6-9, 2019}. OpenReview.net.

\bibitem[{Frankle et~al.(2020)Frankle, Dziugaite, Roy, and Carbin}]{DBLP:conf/icml/FrankleD0C20}
Jonathan Frankle, Gintare~Karolina Dziugaite, Daniel~M. Roy, and Michael Carbin. 2020.
\newblock \href {http://proceedings.mlr.press/v119/frankle20a.html} {Linear mode connectivity and the lottery ticket hypothesis}.
\newblock In \emph{Proceedings of the 37th International Conference on Machine Learning, {ICML} 2020, 13-18 July 2020, Virtual Event}, volume 119 of \emph{Proceedings of Machine Learning Research}, pages 3259--3269. {PMLR}.

\bibitem[{Gao et~al.(2021)Gao, Yao, and Chen}]{DBLP:conf/emnlp/GaoYC21}
Tianyu Gao, Xingcheng Yao, and Danqi Chen. 2021.
\newblock \href {https://doi.org/10.18653/v1/2021.emnlp-main.552} {Simcse: Simple contrastive learning of sentence embeddings}.
\newblock In \emph{Proceedings of the 2021 Conference on Empirical Methods in Natural Language Processing, {EMNLP} 2021, Virtual Event / Punta Cana, Dominican Republic, 7-11 November, 2021}, pages 6894--6910. Association for Computational Linguistics.

\bibitem[{Hill et~al.(2016)Hill, Cho, and Korhonen}]{hill-etal-2016-learning}
Felix Hill, Kyunghyun Cho, and Anna Korhonen. 2016.
\newblock \href {https://doi.org/10.18653/v1/N16-1162} {Learning distributed representations of sentences from unlabelled data}.
\newblock In \emph{Proceedings of the 2016 Conference of the North {A}merican Chapter of the Association for Computational Linguistics: Human Language Technologies}, pages 1367--1377, San Diego, California. Association for Computational Linguistics.

\bibitem[{Hou et~al.(2020)Hou, Huang, Shang, Jiang, Chen, and Liu}]{DBLP:conf/nips/HouHSJCL20}
Lu~Hou, Zhiqi Huang, Lifeng Shang, Xin Jiang, Xiao Chen, and Qun Liu. 2020.
\newblock \href {https://proceedings.neurips.cc/paper/2020/hash/6f5216f8d89b086c18298e043bfe48ed-Abstract.html} {Dynabert: Dynamic {BERT} with adaptive width and depth}.
\newblock In \emph{Advances in Neural Information Processing Systems 33: Annual Conference on Neural Information Processing Systems 2020, NeurIPS 2020, December 6-12, 2020, virtual}.

\bibitem[{Kiros et~al.(2015)Kiros, Zhu, Salakhutdinov, Zemel, Urtasun, Torralba, and Fidler}]{kiros2015skip}
Ryan Kiros, Yukun Zhu, Russ~R Salakhutdinov, Richard Zemel, Raquel Urtasun, Antonio Torralba, and Sanja Fidler. 2015.
\newblock Skip-thought vectors.
\newblock \emph{Advances in neural information processing systems}, 28.

\bibitem[{Li et~al.(2020{\natexlab{a}})Li, Zhou, He, Wang, Yang, and Li}]{DBLP:conf/emnlp/LiZHWYL20}
Bohan Li, Hao Zhou, Junxian He, Mingxuan Wang, Yiming Yang, and Lei Li. 2020{\natexlab{a}}.
\newblock \href {https://doi.org/10.18653/v1/2020.emnlp-main.733} {On the sentence embeddings from pre-trained language models}.
\newblock In \emph{Proceedings of the 2020 Conference on Empirical Methods in Natural Language Processing, {EMNLP} 2020, Online, November 16-20, 2020}, pages 9119--9130. Association for Computational Linguistics.

\bibitem[{Li et~al.(2020{\natexlab{b}})Li, Zhou, He, Wang, Yang, and Li}]{li-etal-2020-sentence}
Bohan Li, Hao Zhou, Junxian He, Mingxuan Wang, Yiming Yang, and Lei Li. 2020{\natexlab{b}}.
\newblock \href {https://doi.org/10.18653/v1/2020.emnlp-main.733} {On the sentence embeddings from pre-trained language models}.
\newblock In \emph{Proceedings of the 2020 Conference on Empirical Methods in Natural Language Processing (EMNLP)}, pages 9119--9130, Online. Association for Computational Linguistics.

\bibitem[{Lin et~al.(2017)Lin, Feng, dos Santos, Yu, Xiang, Zhou, and Bengio}]{DBLP:conf/iclr/LinFSYXZB17}
Zhouhan Lin, Minwei Feng, C{\'{\i}}cero~Nogueira dos Santos, Mo~Yu, Bing Xiang, Bowen Zhou, and Yoshua Bengio. 2017.
\newblock \href {https://openreview.net/forum?id=BJC\_jUqxe} {A structured self-attentive sentence embedding}.
\newblock In \emph{5th International Conference on Learning Representations, {ICLR} 2017, Toulon, France, April 24-26, 2017, Conference Track Proceedings}. OpenReview.net.

\bibitem[{Liu et~al.(2021)Liu, Vulic, Korhonen, and Collier}]{DBLP:conf/emnlp/0001VKC21}
Fangyu Liu, Ivan Vulic, Anna Korhonen, and Nigel Collier. 2021.
\newblock \href {https://doi.org/10.18653/v1/2021.emnlp-main.109} {Fast, effective, and self-supervised: Transforming masked language models into universal lexical and sentence encoders}.
\newblock In \emph{Proceedings of the 2021 Conference on Empirical Methods in Natural Language Processing, {EMNLP} 2021, Virtual Event / Punta Cana, Dominican Republic, 7-11 November, 2021}, pages 1442--1459. Association for Computational Linguistics.

\bibitem[{Liu et~al.(2019)Liu, Ott, Goyal, Du, Joshi, Chen, Levy, Lewis, Zettlemoyer, and Stoyanov}]{DBLP:journals/corr/abs-1907-11692}
Yinhan Liu, Myle Ott, Naman Goyal, Jingfei Du, Mandar Joshi, Danqi Chen, Omer Levy, Mike Lewis, Luke Zettlemoyer, and Veselin Stoyanov. 2019.
\newblock \href {http://arxiv.org/abs/1907.11692} {Roberta: {A} robustly optimized {BERT} pretraining approach}.
\newblock \emph{CoRR}, abs/1907.11692.

\bibitem[{Logeswaran and Lee(2018)}]{DBLP:conf/iclr/LogeswaranL18}
Lajanugen Logeswaran and Honglak Lee. 2018.
\newblock \href {https://openreview.net/forum?id=rJvJXZb0W} {An efficient framework for learning sentence representations}.
\newblock In \emph{6th International Conference on Learning Representations, {ICLR} 2018, Vancouver, BC, Canada, April 30 - May 3, 2018, Conference Track Proceedings}. OpenReview.net.

\bibitem[{Marelli et~al.(2014)Marelli, Menini, Baroni, Bentivogli, Bernardi, and Zamparelli}]{DBLP:conf/lrec/MarelliMBBBZ14}
Marco Marelli, Stefano Menini, Marco Baroni, Luisa Bentivogli, Raffaella Bernardi, and Roberto Zamparelli. 2014.
\newblock \href {http://www.lrec-conf.org/proceedings/lrec2014/summaries/363.html} {A {SICK} cure for the evaluation of compositional distributional semantic models}.
\newblock In \emph{Proceedings of the Ninth International Conference on Language Resources and Evaluation, {LREC} 2014, Reykjavik, Iceland, May 26-31, 2014}, pages 216--223. European Language Resources Association {(ELRA)}.

\bibitem[{Michel et~al.(2019)Michel, Levy, and Neubig}]{DBLP:conf/nips/MichelLN19}
Paul Michel, Omer Levy, and Graham Neubig. 2019.
\newblock \href {https://proceedings.neurips.cc/paper/2019/hash/2c601ad9d2ff9bc8b282670cdd54f69f-Abstract.html} {Are sixteen heads really better than one?}
\newblock In \emph{Advances in Neural Information Processing Systems 32: Annual Conference on Neural Information Processing Systems 2019, NeurIPS 2019, December 8-14, 2019, Vancouver, BC, Canada}, pages 14014--14024.

\bibitem[{Molchanov et~al.(2017)Molchanov, Tyree, Karras, Aila, and Kautz}]{DBLP:conf/iclr/MolchanovTKAK17}
Pavlo Molchanov, Stephen Tyree, Tero Karras, Timo Aila, and Jan Kautz. 2017.
\newblock \href {https://openreview.net/forum?id=SJGCiw5gl} {Pruning convolutional neural networks for resource efficient inference}.
\newblock In \emph{5th International Conference on Learning Representations, {ICLR} 2017, Toulon, France, April 24-26, 2017, Conference Track Proceedings}. OpenReview.net.

\bibitem[{Pang and Lee(2004)}]{DBLP:conf/acl/PangL04}
Bo~Pang and Lillian Lee. 2004.
\newblock \href {https://doi.org/10.3115/1218955.1218990} {A sentimental education: Sentiment analysis using subjectivity summarization based on minimum cuts}.
\newblock In \emph{Proceedings of the 42nd Annual Meeting of the Association for Computational Linguistics, 21-26 July, 2004, Barcelona, Spain}, pages 271--278. {ACL}.

\bibitem[{Pang and Lee(2005)}]{DBLP:conf/acl/PangL05}
Bo~Pang and Lillian Lee. 2005.
\newblock \href {https://doi.org/10.3115/1219840.1219855} {Seeing stars: Exploiting class relationships for sentiment categorization with respect to rating scales}.
\newblock In \emph{{ACL} 2005, 43rd Annual Meeting of the Association for Computational Linguistics, Proceedings of the Conference, 25-30 June 2005, University of Michigan, {USA}}, pages 115--124. The Association for Computer Linguistics.

\bibitem[{Prasanna et~al.(2020)Prasanna, Rogers, and Rumshisky}]{prasanna-etal-2020-bert}
Sai Prasanna, Anna Rogers, and Anna Rumshisky. 2020.
\newblock \href {https://doi.org/10.18653/v1/2020.emnlp-main.259} {{W}hen {BERT} {P}lays the {L}ottery, {A}ll {T}ickets {A}re {W}inning}.
\newblock In \emph{Proceedings of the 2020 Conference on Empirical Methods in Natural Language Processing (EMNLP)}, pages 3208--3229, Online. Association for Computational Linguistics.

\bibitem[{Qiu et~al.(2022)Qiu, Huang, Yin, and Wang}]{DBLP:conf/wsdm/QiuHYW22}
Ruihong Qiu, Zi~Huang, Hongzhi Yin, and Zijian Wang. 2022.
\newblock \href {https://doi.org/10.1145/3488560.3498433} {Contrastive learning for representation degeneration problem in sequential recommendation}.
\newblock In \emph{{WSDM} '22: The Fifteenth {ACM} International Conference on Web Search and Data Mining, Virtual Event / Tempe, AZ, USA, February 21 - 25, 2022}, pages 813--823. {ACM}.

\bibitem[{Reimers et~al.(2016)Reimers, Beyer, and Gurevych}]{reimers-etal-2016-task}
Nils Reimers, Philip Beyer, and Iryna Gurevych. 2016.
\newblock \href {https://aclanthology.org/C16-1009} {Task-oriented intrinsic evaluation of semantic textual similarity}.
\newblock In \emph{Proceedings of {COLING} 2016, the 26th International Conference on Computational Linguistics: Technical Papers}, pages 87--96, Osaka, Japan. The COLING 2016 Organizing Committee.

\bibitem[{Reimers and Gurevych(2019)}]{reimers-gurevych-2019-sentence}
Nils Reimers and Iryna Gurevych. 2019.
\newblock \href {https://doi.org/10.18653/v1/D19-1410} {Sentence-{BERT}: Sentence embeddings using {S}iamese {BERT}-networks}.
\newblock In \emph{Proceedings of the 2019 Conference on Empirical Methods in Natural Language Processing and the 9th International Joint Conference on Natural Language Processing (EMNLP-IJCNLP)}, pages 3982--3992, Hong Kong, China. Association for Computational Linguistics.

\bibitem[{Socher et~al.(2013)Socher, Perelygin, Wu, Chuang, Manning, Ng, and Potts}]{DBLP:conf/emnlp/SocherPWCMNP13}
Richard Socher, Alex Perelygin, Jean Wu, Jason Chuang, Christopher~D. Manning, Andrew~Y. Ng, and Christopher Potts. 2013.
\newblock \href {https://aclanthology.org/D13-1170/} {Recursive deep models for semantic compositionality over a sentiment treebank}.
\newblock In \emph{Proceedings of the 2013 Conference on Empirical Methods in Natural Language Processing, {EMNLP} 2013, 18-21 October 2013, Grand Hyatt Seattle, Seattle, Washington, USA, {A} meeting of SIGDAT, a Special Interest Group of the {ACL}}, pages 1631--1642. {ACL}.

\bibitem[{Su et~al.(2021)Su, Cao, Liu, and Ou}]{DBLP:journals/corr/abs-2103-15316}
Jianlin Su, Jiarun Cao, Weijie Liu, and Yangyiwen Ou. 2021.
\newblock \href {http://arxiv.org/abs/2103.15316} {Whitening sentence representations for better semantics and faster retrieval}.
\newblock \emph{CoRR}, abs/2103.15316.

\bibitem[{Voorhees and Tice(2000)}]{DBLP:conf/sigir/VoorheesT00}
Ellen~M. Voorhees and Dawn~M. Tice. 2000.
\newblock \href {https://doi.org/10.1145/345508.345577} {Building a question answering test collection}.
\newblock In \emph{{SIGIR} 2000: Proceedings of the 23rd Annual International {ACM} {SIGIR} Conference on Research and Development in Information Retrieval, July 24-28, 2000, Athens, Greece}, pages 200--207. {ACM}.

\bibitem[{Wang and Isola(2020)}]{DBLP:conf/icml/0001I20}
Tongzhou Wang and Phillip Isola. 2020.
\newblock \href {http://proceedings.mlr.press/v119/wang20k.html} {Understanding contrastive representation learning through alignment and uniformity on the hypersphere}.
\newblock In \emph{Proceedings of the 37th International Conference on Machine Learning, {ICML} 2020, 13-18 July 2020, Virtual Event}, volume 119 of \emph{Proceedings of Machine Learning Research}, pages 9929--9939. {PMLR}.

\bibitem[{Wiebe et~al.(2005)Wiebe, Wilson, and Cardie}]{DBLP:journals/lre/WiebeWC05}
Janyce Wiebe, Theresa Wilson, and Claire Cardie. 2005.
\newblock \href {https://doi.org/10.1007/s10579-005-7880-9} {Annotating expressions of opinions and emotions in language}.
\newblock \emph{Lang. Resour. Evaluation}, 39(2-3):165--210.

\bibitem[{Xia et~al.(2022)Xia, Zhong, and Chen}]{DBLP:conf/acl/XiaZC22}
Mengzhou Xia, Zexuan Zhong, and Danqi Chen. 2022.
\newblock \href {https://doi.org/10.18653/v1/2022.acl-long.107} {Structured pruning learns compact and accurate models}.
\newblock In \emph{Proceedings of the 60th Annual Meeting of the Association for Computational Linguistics (Volume 1: Long Papers), {ACL} 2022, Dublin, Ireland, May 22-27, 2022}, pages 1513--1528. Association for Computational Linguistics.

\bibitem[{Yan et~al.(2021)Yan, Li, Wang, Zhang, Wu, and Xu}]{DBLP:conf/acl/YanLWZWX20}
Yuanmeng Yan, Rumei Li, Sirui Wang, Fuzheng Zhang, Wei Wu, and Weiran Xu. 2021.
\newblock \href {https://doi.org/10.18653/v1/2021.acl-long.393} {Consert: {A} contrastive framework for self-supervised sentence representation transfer}.
\newblock In \emph{Proceedings of the 59th Annual Meeting of the Association for Computational Linguistics and the 11th International Joint Conference on Natural Language Processing, {ACL/IJCNLP} 2021, (Volume 1: Long Papers), Virtual Event, August 1-6, 2021}, pages 5065--5075. Association for Computational Linguistics.

\bibitem[{Yang et~al.(2022{\natexlab{a}})Yang, Zhang, and Song}]{DBLP:conf/emnlp/YangZS22}
Yi~Yang, Chen Zhang, and Dawei Song. 2022{\natexlab{a}}.
\newblock \href {https://aclanthology.org/2022.emnlp-main.258} {Sparse teachers can be dense with knowledge}.
\newblock In \emph{Proceedings of the 2022 Conference on Empirical Methods in Natural Language Processing, {EMNLP} 2022, Abu Dhabi, United Arab Emirates, December 7-11, 2022}, pages 3904--3915. Association for Computational Linguistics.

\bibitem[{Yang et~al.(2022{\natexlab{b}})Yang, Zhang, Wang, and Song}]{DBLP:conf/nlpcc/YangZWS22}
Yi~Yang, Chen Zhang, Benyou Wang, and Dawei Song. 2022{\natexlab{b}}.
\newblock \href {https://doi.org/10.1007/978-3-031-17120-8\_12} {Doge tickets: Uncovering domain-general language models by playing lottery tickets}.
\newblock In \emph{Natural Language Processing and Chinese Computing - 11th {CCF} International Conference, {NLPCC} 2022, Guilin, China, September 24-25, 2022, Proceedings, Part {I}}, volume 13551 of \emph{Lecture Notes in Computer Science}, pages 144--156. Springer.

\bibitem[{Zhang et~al.(2020{\natexlab{a}})Zhang, He, Liu, Lim, and Bing}]{DBLP:conf/emnlp/ZhangHLLB20}
Yan Zhang, Ruidan He, Zuozhu Liu, Kwan~Hui Lim, and Lidong Bing. 2020{\natexlab{a}}.
\newblock \href {https://doi.org/10.18653/v1/2020.emnlp-main.124} {An unsupervised sentence embedding method by mutual information maximization}.
\newblock In \emph{Proceedings of the 2020 Conference on Empirical Methods in Natural Language Processing, {EMNLP} 2020, Online, November 16-20, 2020}, pages 1601--1610. Association for Computational Linguistics.

\bibitem[{Zhang et~al.(2020{\natexlab{b}})Zhang, He, Liu, Lim, and Bing}]{zhang-etal-2020-unsupervised}
Yan Zhang, Ruidan He, Zuozhu Liu, Kwan~Hui Lim, and Lidong Bing. 2020{\natexlab{b}}.
\newblock \href {https://doi.org/10.18653/v1/2020.emnlp-main.124} {An unsupervised sentence embedding method by mutual information maximization}.
\newblock In \emph{Proceedings of the 2020 Conference on Empirical Methods in Natural Language Processing (EMNLP)}, pages 1601--1610, Online. Association for Computational Linguistics.

\end{thebibliography}

\clearpage
\appendix


\begin{table*}[ht]
\renewcommand{\arraystretch}{}
\resizebox{\textwidth}{!}{
\begin{tabular}{@{}ccccccccc@{}}
\toprule
\multicolumn{1}{l}{}    
& MR & CR & SUBJ & MPQA	& SST2	& TREC	& MRPC	& Avg \\ \midrule

SimCSE-BERT\textsubscript{base}                  
& 78.84 & 84.21 & 93.83 & 88.87 & 83.75 & 86.40 & 72.99  & 84.13 \\
SparseCSE\textsubscript{2\%} 
& $80.88^{\textcolor{red!70}{+\textbf{2.04}}}$ 
& $86.15^{\textcolor{red!70}{+\textbf{1.94}}}$ 
& $94.29^{\textcolor{red!70}{+\textbf{0.46}}}$ 
& $89.40^{\textcolor{red!70}{+\textbf{0.53}}}$ 
& $84.95^{\textcolor{red!70}{+\textbf{1.20}}}$ 
& $88.40^{\textcolor{red!70}{+\textbf{2.00}}}$ 
& $75.54^{\textcolor{red!70}{+\textbf{2.55}}}$  
& $85.66^{\textcolor{red!70}{+\textbf{1.53}}}$ \\

SparseCSE\textsubscript{best} 
& $80.90^{\textcolor{red!70}{+\textbf{2.06}}}_{3\%}$ 
& $86.15^{\textcolor{red!70}{+\textbf{1.94}}}_{2\%}$ 
& $94.58^{\textcolor{red!70}{+\textbf{0.75}}}_{7\%}$ 
& $89.43^{\textcolor{red!70}{+\textbf{0.56}}}_{4\%}$ 
& $85.83^{\textcolor{red!70}{+\textbf{2.08}}}_{3\%}$ 
& $88.40^{\textcolor{red!70}{+\textbf{2.00}}}_{2\%}$ 
& $76.12^{\textcolor{red!70}{+\textbf{3.13}}}_{8\%}$  
& $85.92^{\textcolor{red!70}{+\textbf{1.79}}}_{}$ \\
\midrule
SimCSE-BERT\textsubscript{large}                  
& 84.02 & 88.11 & 94.8 & 89.59 & 89.9 & 90.20 & 75.48  & 87.44 \\
SparseCSE\textsubscript{2\%}  
& $84.26^{\textcolor{red!70}{+\textbf{0.24}}}$ 
& $89.43^{\textcolor{red!70}{+\textbf{1.32}}}$ 
& $95.27^{\textcolor{red!70}{+\textbf{0.47}}}$  
& $89.83^{\textcolor{red!70}{+\textbf{0.24}}}$  
& $89.57^{\textcolor{red!70}{-\textbf{0.33}}}$  
& $92.40^{\textcolor{red!70}{+\textbf{2.20}}}$   
& $76.46^{\textcolor{red!70}{+\textbf{0.98}}}$   
& $88.17^{\textcolor{red!70}{+\textbf{0.73}}}$  \\
SparseCSE\textsubscript{best}  
& $84.65^{\textcolor{red!70}{+\textbf{0.63}}}_{3\%}$ 
& $89.43^{\textcolor{red!70}{+\textbf{1.32}}}_{2\%}$ 
& $95.27^{\textcolor{red!70}{+\textbf{0.47}}}_{2\%}$  
& $90.07^{\textcolor{red!70}{+\textbf{0.48}}}_{9\%}$  
& $89.57^{\textcolor{red!70}{-\textbf{0.33}}}_{2\%}$  
& $93.80^{\textcolor{red!70}{+\textbf{3.60}}}_{6\%}$   
& $76.52^{\textcolor{red!70}{+\textbf{1.04}}}_{3\%}$   
& $88.44^{\textcolor{red!70}{+\textbf{0.99}}}_{}$  \\

\midrule
SimCSE-Roberta\textsubscript{base}                  
& 81.39 & 86.94 & 93.20 & 87.11 & 87.10 & 84.20 & 74.09  & 84.86 \\
SparseCSE\textsubscript{1\%} 
& $82.18^{\textcolor{red!70}{+\textbf{0.79}}}$ 
& $88.05^{\textcolor{red!70}{+\textbf{1.11}}}$ 
& $93.53^{\textcolor{red!70}{+\textbf{0.33}}}$  
& $87.59^{\textcolor{red!70}{+\textbf{0.48}}}$  
& $87.48^{\textcolor{red!70}{+\textbf{0.38}}}$  
& $84.00^{\textcolor{red!70}{-\textbf{0.20}}}$   
& $74.78^{\textcolor{red!70}{+\textbf{0.69}}}$   
& $85.37^{\textcolor{red!70}{+\textbf{0.51}}}$  \\
SparseCSE\textsubscript{best}  
& $82.18^{\textcolor{red!70}{+\textbf{0.79}}}_{1\%}$ 
& $88.21^{\textcolor{red!70}{+\textbf{1.27}}}_{3\%}$ 
& $93.53^{\textcolor{red!70}{+\textbf{0.33}}}_{1\%}$  
& $87.59^{\textcolor{red!70}{+\textbf{0.48}}}_{1\%}$  
& $87.48^{\textcolor{red!70}{+\textbf{0.38}}}_{1\%}$  
& $86.00^{\textcolor{red!70}{+\textbf{1.80}}}_{7\%}$   
& $74.78^{\textcolor{red!70}{+\textbf{0.69}}}_{1\%}$   
& $85.64^{\textcolor{red!70}{+\textbf{0.78}}}_{}$  \\

\bottomrule
\end{tabular}
}

\caption{The result of transfer learning tasks. Data annotation method is the same as the previous table.}
\label{tab:transfer}
\end{table*}

\section{Implementation Details} \label{Pruning}
We follow the training details of SimCSE~\citep{DBLP:conf/emnlp/GaoYC21} for both training and rewinding process of sparseCSE, including hyperparameter settings and a dataset of one million randomly selected sentences from English Wikipedia. 
 
We prune the baseline models on the dataset STS Benchmark~\citep{cer-etal-2017-semeval}. The dataset was originally used to evaluate the alignment and uniformity of sentence embeddings in SimCSE~\citep{DBLP:conf/emnlp/GaoYC21}, and we find that it can also play a role on calculating pruning scores. 
During the pruning process, we explore different sparsity levels from a predefined set (1\%, 2\%, 3\%, 4\%, 5\%, 6\%, 7\%, 8\%, 9\%, 10\%, 20\%, 30\%, 40\%, 50\%), and use a $\lambda$ value of 0.5 for the main experiment. Additionally, we examine the impact of different $\lambda$ values (0.25 and 0.75) in further analysis.

\begin{figure}[h]
\centering
    \includegraphics[width=0.45\textwidth]{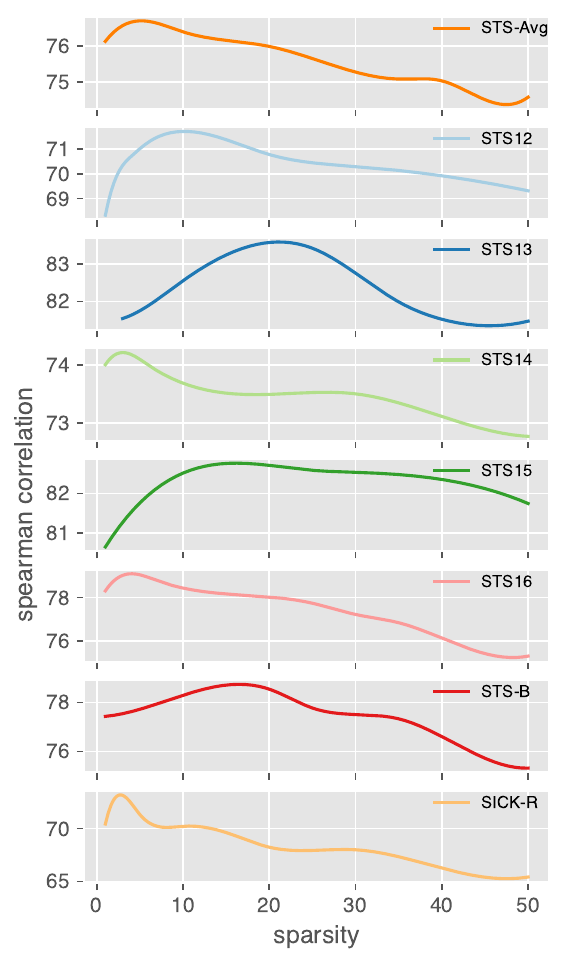}
    \caption{Transitions with varying sparsity on STS tasks.}
    \label{fig:sub_a}
\end{figure}

\begin{figure}[h]
\centering
    \includegraphics[width=0.45\textwidth]{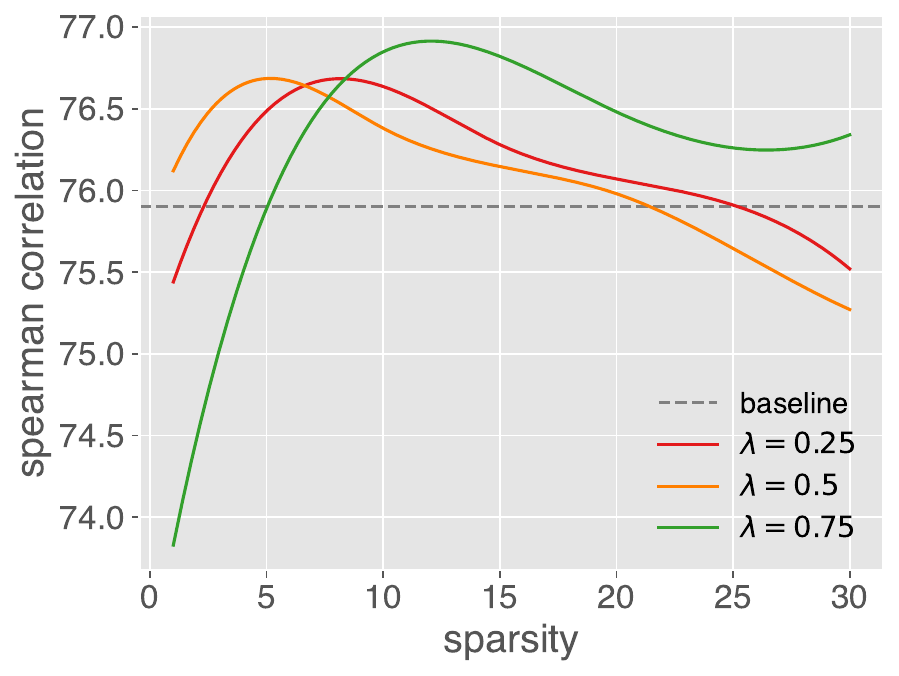}
    \caption{Average STS performance of SparseCSE using $BERT_{base}$ with different $\lambda$.}
    \label{fig:sub_b}
\end{figure}
\section{Transfer Tasks for Evaluation} \label{Transfer}

The transfer learning tasks contain MR~\citep{DBLP:conf/acl/PangL05}, 
CR~\citep{DBLP:conf/sigir/AmplayoBS0L22}, 
SUBJ~\citep{DBLP:conf/acl/PangL04}, 
MPQA~\citep{DBLP:journals/lre/WiebeWC05}, 
SST-2~\citep{DBLP:conf/emnlp/SocherPWCMNP13}, 
TREC~\citep{DBLP:conf/sigir/VoorheesT00} 
and MRPC~\citep{dolan-brockett-2005-automatically}, which are different sentence classification tasks and can give an impression on the quality of sentence embeddings.

The results on transfer learning tasks are shown in table~\ref{tab:transfer}. And the average improvement on BERT\textsubscript{base}, BERT\textsubscript{large} and Roberta\textsubscript{base} achieves 1.79\%, 0.99\% and 0.78\%, respectively.
For instance, when applying 2\% sparsity to the BERT\textsubscript{base} model, we achieve the best average improvement of 1.53 on transfer tasks shown in Table~\ref{tab:transfer}. All tasks benefit from this pruning sparsity, with improvements of 2.04, 1.94, 0.46, 0.53, 1.20, 2.00, and 2.55. 

\section{Searching within Varying Sparsity}
\label{6.3}
The transition of the BERT\textsubscript{base} model's performance, as measured by the average score across the seven STS tasks, as well as the discrete scores of these tasks, is illustrated in Figure \ref{fig:sub_a}. It is evident from the figure that for each task, the model's performance initially improves and then declines as the sparsity level increases, showing a peak. In every task, this peak appears steadily around a fixed sparsity corresponding to the optimal sparsity value in the main results. This indicates that the best performance observed in the main results for each task is not an isolated occurrence but rather a continuous trend.

\section{Tradeoff of Alignment and Uniformity} 
\label{6.4}
In our approach, the alignment loss and uniformity loss work together to guide parameter scoring, with the coefficient $\lambda$ regulating their relative influence.
To further investigate the contributions of alignment and uniformity strategies to model effectiveness, we conducted additional experiments using different $\lambda$ values (0.25, 0.5, 0.75) as shown in Figure~\ref{fig:sub_b}.
We observed that the coefficient does not have a significant impact on the peak value of each task. However, it does influence the pattern of how model performance varies with sparsity.
When $\lambda=0.5$, the pruned model's performance exhibits a rapid increase and decrease at lower sparsity levels, resulting in a distinct peak.
On the other hand, with $\lambda=0.25$, the performance trend shows a relatively flatter increase and decrease, with the peak occurring at slightly higher sparsity levels.
These findings suggest that alignment and uniformity play similar roles in guiding contrastive representation learning, but they have different effects on parameter filtering.

\end{document}